\documentclass[a4paper,fleqn]{cas-sc}

\usepackage{url}
\usepackage{graphicx}
\usepackage{cite}
\usepackage{acronym}
\usepackage{hyperref}
\usepackage{xurl}

\usepackage{acronym} 

\acrodef{SNN}[SNN]{Spiking Neural Network}
\acrodef{STDP}[STDP]{Spike-timing-dependent plasticity}
\acrodef{DG}[DG]{Dentate Gyrus}
\acrodef{PC}[PC]{Pyramidal Cell}
\acrodef{CA}[CA]{Cornu Ammonis}
\acrodef{LIF}[LIF]{Leaky Integrate-and-Fire}
\acrodef{EC}[EC]{Entorhinal Cortex}
\acrodef{ANN}[ANN]{Artificial Neural Network}
\acrodef{LTP}[LTP]{Long-Term Potentiation}
\acrodef{LTD}[LTD]{Long-Term Depression}
\acrodef{ANN}[ANN]{Artificial Neural Network}
\acrodef{CAM}[CAM]{Content-Addressable Memory}

\usepackage[numbers]{natbib}

\def\tsc#1{\csdef{#1}{\textsc{\lowercase{#1}}\xspace}}
\tsc{WGM}
\tsc{QE}

\begin{document}
\let\WriteBookmarks\relax
\def\floatpagepagefraction{1}
\def\textpagefraction{.001}

\shorttitle{A computational approach to a neuromorphic Content-Addressable Memory bio-inspired on the hippocampus}

\shortauthors{D.~Casanueva-Morato, A.~Ayuso-Martinez, J.~P.~Dominguez-Morales, A.~Jimenez-Fernandez, G.~Jimenez-Moreno}  

\title [mode = title]{Bio-inspired computational memory model of the Hippocampus: an approach to a neuromorphic spike-based Content-Addressable Memory}  



%

\affiliation[1]{organization={Escuela Técnica Superior de Ingeniería Informática (ETSII), Universidad de Sevilla}, 
            city={Seville},
            addressline={Avenida de Reina Mercedes s/n},
            postcode={41012},
            country={Spain}}

\affiliation[2]{organization={Robotics and Tech. of Computers Lab., Universidad de Sevilla}, 
            city={Seville},
            postcode={41012},
            country={Spain}}

\affiliation[3]{organization={Escuela Politécnica Superior (EPS), Universidad de Sevilla}, 
            city={Sevilla},
            postcode={41011},
            country={Spain}}

\affiliation[4]{organization={Smart Computer Systems Research and Engineering Lab (SCORE), Research Institute of Computer Engineering (I3US), Universidad de Sevilla}, 
city={Seville},
postcode={41012},
country={Spain}}

\author[1,2,3]{Daniel Casanueva-Morato}[
      orcid=0000-0002-7676-1629,
]
\ead{dcasanueva@us.es}

\author[1,2,3]{Alvaro Ayuso-Martinez}[
      orcid=0000-0002-0059-6647,
]
\ead{aayuso@us.es}

\author[1,2,3,4]{Juan P. Dominguez-Morales}[
      orcid=0000-0002-5474-107X,
]
\ead{jpdominguez@us.es}

\author[1,2,3,4]{Angel Jimenez-Fernandez}[
      orcid=0000-0003-3061-5922,
]
\ead{angel@us.es}

\author[1,2,3,4]{Gabriel Jimenez-Moreno}[
      orcid=0000-0003-4512-6750,
]
\ead{gaji@us.es}



\begin{abstract}
The brain has computational capabilities that surpass those of modern systems, being able to solve complex problems efficiently in a simple way. Neuromorphic engineering aims to mimic biology in order to develop new systems capable of incorporating such capabilities. Bio-inspired learning systems continue to be a challenge that must be solved, and much work needs to be done in this regard. Among all brain regions, the hippocampus stands out as an autoassociative short-term memory with the capacity to learn and recall memories from any fragment of them. These characteristics make the hippocampus an ideal candidate for developing bio-inspired learning systems that, in addition, resemble content-addressable memories. Therefore, in this work we propose a bio-inspired spiking content-addressable memory model based on the CA3 region of the hippocampus with the ability to learn, forget and recall memories, both orthogonal and non-orthogonal, from any fragment of them. The model was implemented on the SpiNNaker hardware platform using Spiking Neural Networks. A set of experiments based on functional, stress and applicability tests were performed to demonstrate its correct functioning. This work presents the first hardware implementation of a fully-functional bio-inspired spiking hippocampal content-addressable memory model, paving the way for the development of future more complex neuromorphic systems.
\end{abstract}



\begin{keywords}
Hippocampus model \sep Content-Addressable memory \sep Spiking Neural Networks \sep Neuromorphic engineering \sep CA3 \sep SpiNNaker
\end{keywords}

\maketitle


\section{Introduction}
\label{sec:introduction}

The evolution of technology and its proliferation as an application in various fields has brought with it the need to develop systems capable of dealing with the large amount of data to be processed and stored \cite{vanarse2019neuromorphic, soman2016recent, zenke2021visualizing}. Given this situation, neuromorphic engineering proposes the study and development of bio-inspired brain architectures to incorporate the power, energy efficiency and computational capabilities observed in nature into existing systems \cite{mead1990neuromorphic, vanarse2019neuromorphic, indiveri2011neuromorphic}. This paradigm has become increasingly important in the last few years \cite{soman2016recent, zenke2021visualizing}.

In general, these systems present an approach based on \acp{SNN}, i.e., networks of artificial neurons, biologically close to those present in biology, interconnected through synapses. In these networks, communication between neurons takes place asynchronously through the generation of action potentials (also called spikes) and learning is achieved through the plasticity of the synapses connecting the different neurons. Thanks to this spiking approach, neuromorphic systems have great advantages in terms of energy consumption and real-time operation compared to traditional systems \cite{zhu2020comprehensiveReview, NeuromorphNature}.

Within the brain, there are different regions responsible for learning and storing the stimuli and information that animals constantly receive. Among all the regions, the hippocampus stands out for its capacity as a short-term associative memory system capable of storing large amounts of information from different cortical brain regions without prior interpretation, without structure and in a rapid manner \cite{rolls2021brain}. This learning is achieved thanks to the recurrent collateral neural network structure present in the Cornu Ammonis 3 (CA3) region within the hippocampus. This network is formed by a series of interconnected pyramidal neurons and interneurons capable of learning information patterns and recalling them from a fragment of them \cite{wible2013hippocampal, rolls2021brain}.

On the one hand, when the information patterns arrive at CA3, oscillatory activity occurs, modifying the synaptic weights of a set of collateral synapses. This process is responsible for learning and storing the input pattern by association of its different fragments. On the other hand, when a fragment of a previously learned pattern arrives, after a series of oscillations regulated by the modified synaptic weights, the network is able to recall the complete pattern \cite{rolls2021brain}. In addition, the pattern can be recalled from different fragments of itself thanks to the topological distribution of the connections between the different neurons and interneurons in this region \cite{wible2013hippocampal, rolls2021brain}. These features make CA3 a biological \ac{CAM} \cite{mueller1999content}.

In recent years, the direction taken in \ac{CAM} system design has focused on ferroelectric transistors technology \cite{ni2019ferroelectric, khan2020future, kazemi2021fefet} and memristors \cite{li2020analog, graves2020memory}. However, studies such as \cite{karam2015emerging} conducted literature reviews of \ac{CAM} systems and mention neuromorphic systems and self-associative networks as an emerging approach in this field. It focuses on proposals based on \acp{ANN} \cite{sivaganesan2020event} and mentions the great potential of \acp{SNN} with the \ac{STDP} learning mechanism for the elaboration of CAM architectures.

In the literature, the spectrum of works addressing spiking \ac{CAM} systems is limited. The main proposals, \cite{mueller1999content, matsugu1994spatiotemporal}, present fully-connected recurrent SNN models as spiking CAM systems. These studies made use of bio-inspired mechanisms, such as oscillatory activity or delayed feedback, to demonstrate the capacity for learning and content-addressable recalling in this type of network. The conclusions drawn from these works indicated the suitability of the CA3 region of the hippocampus for this task, as well as the correct functioning of the network for orthogonal patterns (patterns without overlap), while presenting problems when recalling non-orthogonal patterns (patterns with a certain degree of overlap). These problems were derived from the very definition of non-orthogonal patterns; as there was some overlapping between them, the recall of these patterns was not perfect, since they presented elements of all the intervening patterns.

Regarding bio-inspired spiking memory systems, more authors have addressed this topic. On the one hand, there are authors who proposed hybrid memory systems between \ac{ANN} and \ac{SNN} \cite{zhang2016hmsnn}, or directly used \acp{ANN} and performed a direct translation from \ac{ANN} to \ac{SNN} \cite{yue2023hybrid}. Therefore, they are not purely spiking and are not bio-inspired. On the other hand, authors such as \cite{tan2011associative, tan2013hippocampus, casanueva2022spike} propose purely spiking memory models whose architecture is bio-inspired in the hippocampus and are able to learn and recall patterns. However, their storage capacity is quite small, on the order of a few "bits" of information, and, although they work well for orthogonal patterns, they present problems when dealing with non-orthogonal patterns. In \cite{casanueva2022bio}, a bio-inspired hippocampal memory model is proposed, which is capable of learning and recalling both orthogonal and non-orthogonal patterns, but only from a single fragment of itself called cue.

There are alternative approaches to the topic, such as \cite{shiva2016continuous}, which introduces a theoretical and mathematical model of a bio-inspired memory based on the CA3 region of the hippocampus, although there are no details about its functioning or characteristics. In \cite{he2019constructing}, the authors proposed an associative memory that has a synapse pruning phase, sacrificing the plasticity and dynamism of the synapses themselves.

In short, the characteristics of the hippocampus propose it as a potential solution to the problem of the exponential increase in information processing and storage. However, references and proposals for this type of spiking systems are scarce and limited in the literature. In view of this, this paper proposes a spiking CAM memory model bioinspired in the CA3 region of the hippocampus. The main contributions of this work include the following:
\begin{itemize}
    \item A dynamic bio-inspired spiking memory model based on the hippocampus capable of learning patterns and recalling them, not only from the cue that identifies it, but also from any part of its content.
    \item Implementation of the proposed SNN model on the SpiNNaker hardware platform.
    \item The source code is publicly available, together with the documentation including all the necessary details regarding the \ac{SNN} architectures.
\end{itemize}

The rest of the paper is structured as follows: Section~\ref{sec:materials} presents the materials used in this work. In Section~\ref{sec:model}, the proposed model is detailed, including its architecture (Section~\ref{subsec:architecture}) and its operation principle (Section~\ref{subsec:operating_principle}). The experiments performed to evaluate the functionality and performance of the proposed model are explained in Section~\ref{sec:experiments_and_results}, along with the results obtained. Then, in Section~\ref{sec:discussion}, the results of the experiments are discussed. Finally, the conclusions of the paper are presented in Section~\ref{sec:conclusions}.

\section{Materials}
\label{sec:materials}

When working with neuromorphic systems, SNNs are commonly used. These networks are based on the use of a set of elements (neurons, synapses and learning rules) that not only are inspired by biology but aim to mimic it in order to incorporate the great neurocomputational capabilities observed in nature \cite{ahmed2020brain}.

The principle of operation and communication of these networks of neurons is the generation and propagation of asynchronous electrical pulses or spikes. This makes them very efficient from a computational point of view, as they only operate when an event occurs (when they generate and/or receive a spike) and only in those parts of the network that are affected by the event \cite{tavanaei2019deep}.

These networks are made up of a set of neurons interconnected by synapses. These neurons have an internal state, i.e., a membrane potential, which increases with the arrival of current (spikes) from other neurons connected by synapses. When the membrane potential reaches a certain threshold, the neuron generates a spike. Of all the neuron models, the one most widely used in the literature is the \ac{LIF} \cite{stein1965theoretical, tavanaei2019deep}. Synapses are modelled as connections with static or dynamic weights and a time delay.

Learning in SNNs is generally based on the plasticity of synapses, i.e., the ability to modify the weight of synapses. Of all the existing mechanisms, the \ac{STDP} learning mechanism stands out. This is a Hebbian learning mechanism in which the weight of synapses is modified in proportion to the degree of temporal correlation between the activity of pre- and post-synaptic neurons \cite{sjostrom2010spike}. Moreover, this learning mechanism is distributed, obtaining the necessary information from the spikes encoded between neurons to define the weight of the synapses \cite{caporale2008spike, ahmed2020brain}.

\acp{SNN} present a set of characteristics and features that favor their implementation in hardware platforms in comparison to traditional neural networks \cite{lobo2020spiking}. Different hardware platforms particularly designed for implementing and simulating \acp{SNN} can be found in the literature \cite{furber2014spinnaker, davies2018loihi, merolla2014million}. In this work, we used SpiNNaker \cite{furber2014spinnaker}. This platform works with a standard time step of 1~millisecond (ms). Furthermore, the implementation of the STDP learning mechanism has a residual value greater than 0 even when sufficient time has elapsed between post- and pre-synaptic spikes \cite{jin2010implementing}. This causes a constant degeneration in the weights of the dynamic synapses with each operation.

\section{Hippocampus computational memory model}
\label{sec:model}

\subsection{Architecture}
\label{subsec:architecture}

\begin{figure*}[!t]
    \centering
    \includegraphics[width=0.5\textwidth]{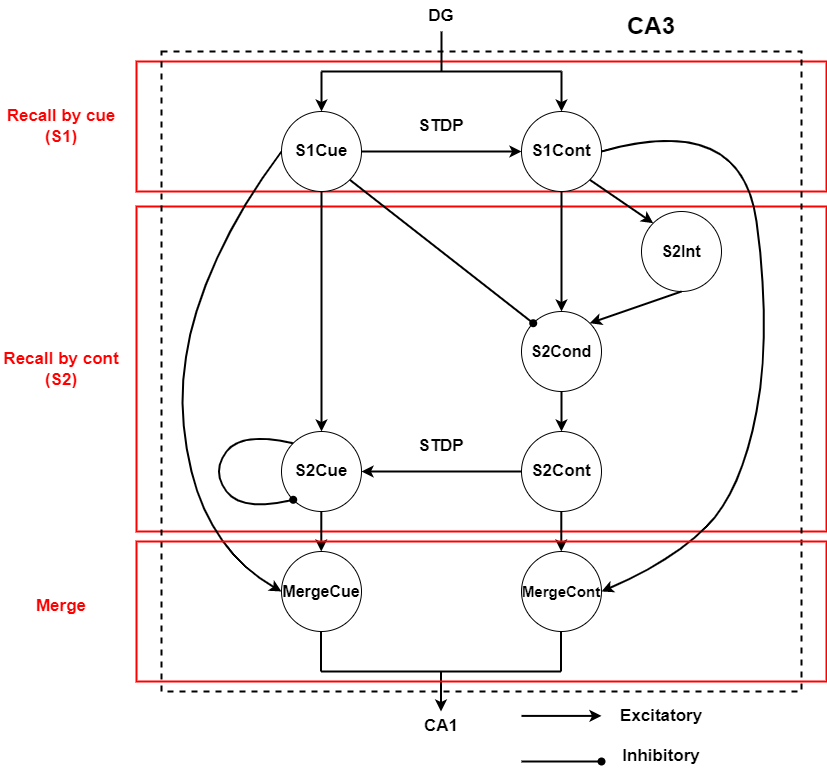}
    \caption{Architecture of the proposed CAM model bioinspired in the CA3 region of the hippocampus. Each circle represents a subpopulation of pyramidal neurons or interneurons and the arrows refer to the synapses (inhibitory and excitatory) between two subpopulations. The word STDP marks those synapses that exhibit the STDP learning mechanism. The architecture is divided into 3 neuron structures highlighted in red: Recall by cue (S1), Recall by cont (S2) and Merge.}
    \label{fig:memoryarch}
\end{figure*}

The architecture of the proposed CAM model, which is inspired by the CA3 region of the hippocampus, is presented in Figure~\ref{fig:memoryarch}. This model bases its functionality on the proposed distribution of excitatory and inhibitory connections, i.e., the topology of synapses between the different pyramidal neurons and interneurons that form CA3.

The proposed model adopts the nomenclature of CAM memories, thus the memory to be learned and recalled is divided into two parts: a fragment of the memory from which to recall the rest of the memory, also called cue, and the remaining part of the memory recalled through the cue, also called content. Following this, the proposed connectivity topology divides CA3 neurons and interneurons into 3 types of subpopulations.

The 3 types of subpopulations are: cue, content and interneurons. The interneuron subpopulations (S2Int and S2Cond) are in charge of regulating the activity of the network during the different operations of the system (learning, forgetting, recall by cue and recall by content). The subpopulations of pyramidal cue neurons (S1Cue, S2Cue and MergeCue) are in charge of working with spiking activity related to the cue, while the subpopulations of pyramidal neurons related to the content (S1Cont, S2Cont and MergeCont) are in charge of working with spiking activity related to the content.

The proposed distribution of neural connections, and thus of subpopulations, can be divided into 2 main structures. The first structure is in charge of managing the addressing of the memory through the cue (S1), while the second structure is in charge of managing the addressing of the memory through the content (S2). The output activity flow of both neural structures is merged by means of a third neural structure, to equalize the size of the inputs and outputs of the network.

\subsubsection{Recall by cue structure (S1)}
\label{subsubsec:recall_cue}

This structure is based on the CA3 model proposed in \cite{casanueva2022bio}, a previous work that presents a bio-inspired spiking memory architecture based on the hippocampus with large storage capacity and capable of learning and recalling orthogonal and non-orthogonal information patterns from a specific cue.

The proposed CA3 region receives the information flow of the memory separated into two parts: cue and content. This is based on the biological idea that the input to CA3 is modified by \ac{DG}. DG is responsible for increasing the sparsity of the information in the input memory. As the degree of dispersion achieved is not known, it is considered to be the maximum possible, i.e., a one-hot encoding. However, this encoding is only applied to one part of the memory (the cue), while the rest of the memory (the content) remains with the same encoding as the input.

Following this logic, the S1Cue subpopulation expects spiking information related to the cue and therefore encoded in one-hot, i.e., a neuronal activity defined by the activation of at most one neuron at each instant of time. The S1Cont subpopulation, on the other hand, receives the neuronal activity related to the content of the memory.

The basis of this structure lies in the all-to-all STDP learning mechanism synapses from S1Cue to S1Cont. When a memory arrives at the input, the spiking activity reaches both subpopulations at the same instant. This triggers the activation of the STDP mechanism, learning the input memory by association of the activity of both populations and, at the same time, forgetting previously learned memories using the same cue. If the activity of an incoming memory fragment corresponds only to the cue, the rest of the memory will be recalled. However, if the input memory fragment corresponds to the content, it does not generate any response. Therefore, its functionality is focused on learning the association of the cues with the corresponding content.

In addition, forgetting a memory occurs as a consequence of an STDP-induced decrease in weight at those synapses that represent the associations to be forgotten. For this decrease to occur, the temporal activation ratio of the neurons connecting the synapses to be forgotten must be first postsynaptic and then presynaptic. This occurs implicitly during the learning of a memory whose cue is similar to that of a previously learned memory. When learning a new memory, there is also the recall of the previous memory with the same cue. Both operations occur in such a temporal sequence that learning of the new memory and forgetting of the old memory takes place.

\subsubsection{Recall by content structure}
\label{subsubsec:recall_cont}

It complements the previous structure by focusing on learning the associations between specific content with their corresponding cues. For this purpose, it presents the neural subpopulations S2Cue and S2Cont, which are in charge of working with the spiking information related to the cue and the content of the memory, respectively. These subpopulations are connected by synapses with the STDP learning mechanism, but, unlike the previous structure, the direction of these connections is from S2Cont to S2Cue.

When an input memory is received, the spiking activity will reach both populations and the STDP will be activated, storing the input memory. When a memory fragment corresponding to the cue arrives, this structure does not generate a response. However, when a memory fragment corresponding to a part of the content arrives, it generates output activity corresponding to those cue neurons that contain at least part of the input activity pattern in their associated content. In other words, it recalls all cues belonging to memories whose content is partially similar to the input activity. This process is called recall by content in CAM memories.

For the correct functioning of the different operations of the system, a series of changes are introduced with respect to the previous structure. On the one hand, the input of the memory to this structure comes from the first structure and not from an external source. This ensures that, during the forgetting operation, all the content neurons involved in the previously learned memory are activated. In other words, it ensures that all synapses storing the association relations of the previous pattern are affected by the decreasing weight of the STDP mechanism and, thus, that these associations and the pattern are also forgotten in this second structure. If the spiking activity were to come directly from the external source and not from the previous structure, the content neurons to be forgotten that did not belong to the new memory would not be activated, thus the previous memory would not be completely forgotten.

On the other hand, two subpopulations of interneurons, S2Int and S2Cond, are added. These are in charge of regulating the input activity to the S2Cont subpopulation. When a recall by cue takes place in the model, due to the parallel connection of both structures, the cue would reach it first and then the rest of the content of the memory recalled by the first structure. Given the temporal sequence of the STDP mechanism, this would cause the pattern to be forgotten in the present structure. To prevent this, subpopulations of interneurons act as a gateway or inhibitor of activity to S2Cont.

S2Cond receives excitatory synapses from S1Cont and connects exitatorily to S2Cont (both with 1-to-1 connectivity), propagating the activity of the content of the former structure to the latter. However, this activity can be inhibited by the activity of the all-to-all inhibitory synapses from S1Cue with a longer time delay. This temporal difference allows the activity to pass from S1Cont to S2Cont only when the cue and the content of the memory arrive at the same time at the first structure (learning) or when the input to the network consists only of activity that encodes the content (recall by content), whereas it inhibits this activity when the activity occurs first in S1Cue rather than S1Cont (recall by cue).

During the learning process, the memory to be learned is introduced in the input for 3 time steps. The first time step of activity will generate the inhibition that prevents the passage of activity from S1Cont to S2Cont. Therefore, to avoid this unwanted inhibition during learning, the S2Int interneuron is added. This interneuron is activated by the activation of any S1Cont neuron and is connected in an excitatory manner to all S2Cond neurons to counteract the inhibition of S1Cue. However, thanks to the temporal difference between the two inputs, inhibition is only counteracted for learning operations (inhibition and excitation arrive at the same time), while this inhibition is maintained for recall by cue operations (inhibition arrives before excitation).

Finally, inhibitory recurrent all-to-all collateral synapses are added to the S2Cue subpopulation. These synapses prevent recall by content being activated during learning and forgetting. This recall by content operation would trigger the activation of neurons corresponding to the cues of memories other than the ones to be learned and forgotten with a temporal sequence that would lead to the forgetting of parts of these unwanted memories. By avoiding recall by content in this situation, memory leakage is prevented.

\subsubsection{Merge structure}
\label{subsubsec:merge}

It is the structure in charge of merging the flow of activity from both previous structures by the sum of the spiking activity of each output neuron. Therefore, its main function is to maintain a consistent spiking representation of the input memory with respect to the output memory.

\subsection{Operating principle}
\label{subsec:operating_principle}

The proposed model is able to perform the following operations: learning, recall by cue, recall by content and forgetting. When these operations are carried out, it is necessary to take into account the time difference of one operation and the next one. To avoid the result of the STDP mechanism for one operation to be affected by the next operation and vice versa, it is necessary to leave a temporal distance of at least 4 time units (for the configuration used) between the activation of the STDP mechanism for the first operation and the activation of the STDP mechanism for the second operation. An activation of the mechanism with a shorter time distance as a consequence of different operations will result in a degradation of the stored content.

The spiking activity of the model presents a spatial coding in time. This encoding is based on the fact that all neuronal activity within a subpopulation at the same instant of time refers to the same memory, while neuronal activity in the same subpopulation but at different instants of time refers to different memories.

\subsubsection{Learning}
\label{subsubsec:learn}

The learning of a memory occurs due to the association of the neuronal activity that encodes it. When the memory arrives to the network, those synapses with an STDP mechanism whose presynaptic and postsynaptic neurons are activated at the same time will obtain a weight increase. In other words, memories are dynamically stored in the weights of the synapses. 

In addition, the learning process is distributed across the network. On the one hand, the model allows learning by the association of the spiking activity encoding the cue with that encoding the content at synapses connecting the S1Cue subpopulation with S1Cont subpopulation. On the other hand, the model allows learning by the association of the spiking activity encoding the content with that encoding the cue at synapses connecting S2Cont with S2Cue. The set of associations learned in both directions by both structures represent the complete memory and will, subsequently, allow the memory to be recalled by both parts.

The learning operation starts with the arrival of a complete memory (cue and content) encoded in spikes to the network. This activity reaches the S1Cue and S1Cont subpopulations at the same time, triggering their activation and generating a pattern of output activity that is identical to the input. This activity will propagate to the S2Cue and S2Cont subpopulations, which will also be activated and generate an identical output pattern. Finally, all this activity will reach the third structure, where it will be merged together, resulting in an output activity pattern that is identical to the input. During this process, the STDP mechanism will be activated in both structures, i.e., the pattern will be learned by association of its parts in both directions.

In order for the increase in weight given by STDP to be sufficient to learn the memory correctly, it is necessary for this activation to be repeated twice. However, because of the refractory time of the neurons, they cannot be activated twice in a row; a difference of a single time stamp is necessary between two consecutive activations. Therefore, in order to perform a learning operation in the network, it is necessary to input the memory for 3 consecutive time steps. 

This operation has a time cost of 7 time steps from the time the memory arrives until an output is obtained on the network. In addition, after starting a learning operation, 7 time steps are required until the start of the next operation. If  these time constraints are not respected, there would be an interference between the two operations and a consequent degradation of the memory content. Thus, if a learning operation starts at time step 0, the result of the operation would be available at time step 7 and the next operation could start.

\subsubsection{Recall}
\label{subsubsec:recall}

The model presents two ways of recalling or addressing a memory: by cue or by content. On the one hand, recalling a memory by its cue begins by passing a fragment of the memory to the network, specifically, that neuronal activity that encodes the cue. This activity will reach S1Cue, triggering the activation of those neurons that represent the cue and propagating to S1Cont through the synapses with STDP. In S1Cont, those neurons that represent the content and that, thanks to the previous learning process, are associated with those cue neurons, will be activated.

This activity will be inhibited by the structure formed by the S1Cue, S2Int and S2Cond subpopulations (Section~\ref{subsubsec:recall_cont}) in the second structure and will arrive directly to the population in charge of merging the information. The output activity from this population will be identical to the input activity, as it only receives activity from a single population. Therefore, the output activity will be first the cue introduced to start the recall by cue operation and, one time step later, the content associated with that cue.

On the other hand, recall by content begins by passing a fragment of memory associated with the content to the network. The first neural structure does not participate in this operation, and the activity reaches S1Cont. The activity propagates directly to the second structure, i.e., from S1Cont to S2Cont, passing the inhibitory structure of the S1Cue, S2Int and S2Cond subpopulations (Section~\ref{subsubsec:recall_cont}).

This activity propagates to S2Cue, activating all those neurons that encode cues belonging to memories that share at least part of the content with the input content. In other words, those memories whose content partly matches the input content are recalled. This activity will pass through the Merge structure and arrive at the output of the network. This activity will be represented by the content used for the recalling operation, together with the cues of the recalled memories (which will take an additional time step to appear at the output).

The recall operation, either by content or by cue, has a time cost of 6 time steps from the arrival of the memory fragment to be recalled until the result of the operation is obtained at the output. Taking into account the temporal restrictions (4 time steps) between activations of the synapses with STDP, after starting a recall operation, 6 time steps are required until starting the next operation. Thus, if a recall operation starts at time step 0, at time step 5 the pattern fragment entered for recall is obtained, at time step 6 the result of the operation is obtained and, at the same instant, the next operation can be started.

\subsubsection{Forget}
\label{subsubsec:forget}

Forgetting is performed implicitly by performing a learning operation on a memory whose cue has previously been associated with another memory. During the learning operation, in those subpopulations connected by synapses with STDP, the cue-content association is learned twice in a row, separated by a single time step. During this time gap, the recall operation of the previous memory will occur in the first structure. 

As the content of the previously learned memory is activated and immediately followed by the cue, an activation of the STDP occurs, reducing the weight of this association in both structures. This reduction has been adjusted so that a single activation is sufficient to trigger the forgetting of these associations and, thus, of the previous memory.

In the case of the second structure, as a consequence of the inhibitory recurrent collateral synapses in S2Cue, recall by content does not occur, as this would involve forgetting memories other than the one being implicitly forgotten. Furthermore, during a recall by cue, only those content neurons associated with the input cue that do not belong to the new memory will be activated in the first structure. The matching ones were activated at the previous time step, thus they would be in the refractory period at the current time instant. This selective recall leads to selective forgetting that only affects the previous memory, not the new one.

\section{Experimentation and results}
\label{sec:experiments_and_results}

A set of incremental experiments were performed to demonstrate the correct operation of the model. In addition, to illustrating its possible use cases, a robotic application experiment for environment mapping was developed.

The experiment was carried out on the proposed model implemented on the SpiNNaker hardware platform, defining a time step of 1~ms.

Each experiment consisted of a temporal simulation of the model in which the behavior of the network is analyzed given an input activity. The results of each experiment are presented graphically by means of rasterplots. The X-axis represents the temporal evolution of the simulation in ms and the Y-axis represents each neuron of the network identified by the subpopulation that it belongs to and the internal ID within the subpopulation. Each point represents a spike fired by the neuron marked by the Y-axis at the time instant marked by the X-axis.

\subsection{Operation tests}
\label{subsec:test_operations}

\begin{figure*}[!t]
    \centering
    \includegraphics[width=0.75\textwidth]{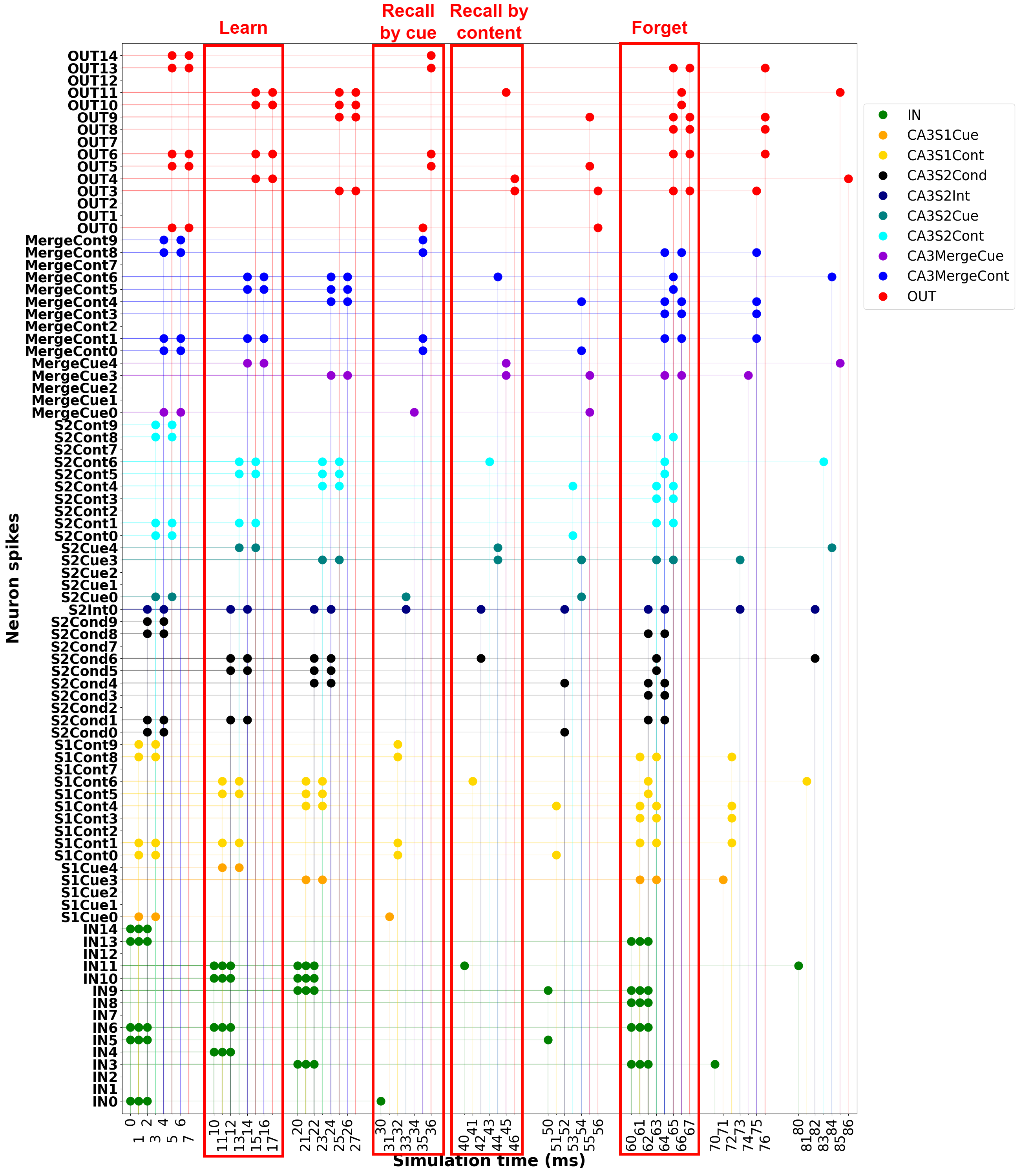}
    \caption{Raster plot of spiking activity of the network during the operation test consisting of learning, recall by content, recall by cue, and forgetting operations.}
    \label{fig:test_operations}
\end{figure*}

This experiment aims to demonstrate the functioning of the different operations of the model. For this purpose, the spiking activity input to the network equivalent to a set of 3 learning, 2 recall by cue, 3 recall by content and 1 learning with implicit forgetting operations were simulated. The simulation has a duration of 86 ms and the result can be seen in Figure~\ref{fig:test_operations}.
 
The model implemented for this experiment is capable of storing a total of 5 memories with a total size of 15 neurons, i.e., equivalent to 15 bits if the information is encoded in binary. Of these 15 neurons, the first 5 encode the cue, while the remaining 10 encode the content. The network architecture is parameterized, therefore, memories of larger or smaller size could be implemented in an easy way. The decision to choose this network size is that the number of neurons and internal activity of the network is small and representative enough to demonstrate the correct functioning of the network while still being able to plot the spiking activity in a clear way.

\subsubsection{Learn}
\label{subsec:test_operations_learn}

The first learning operation starts at ms 0 with a memory formed by the activation of neurons 0, 5, 6, 13 and 14 of the input population for 3  consecutive milliseconds (0, 1 and 2). The activity of the neurons encoding the cue, i.e., the first 5 neurons (0 to 4) arrives to S1Cue, while the activation of the neurons encoding the content, i.e., the next 10 neurons (neurons 5 to 14) arrives to S1Cont. This causes the activation of neuron 0 in S1Cue and neurons 0, 1, 8 and 9 in S1Cont at ms 1 and 3, while at ms 2 the neurons are in the refractory period. This activation of neurons in both subpopulations at the same time triggers the activation of the STDP mechanism and, thus, the learning of the memory from the cue-content associations. 

The neuronal activity propagates, causing the activation of the neuron in S2Int and neurons 0, 1, 8 and 9 in S2Cond at ms 2 and 4. On the one hand, S2Int is activated as long as there is at least one activated neuron in the S1Cont subpopulation. On the other hand, the same neurons are activated in S2Cond as in S1Cont unless it is inhibited by S1Cue neurons. This inhibition, being delayed by 1 ms with respect to the excitatory input of S1Cont (ms 3 and 5), does not prevent its activation. In short, this activity managed to pass the inhibitory structure formed by these subpopulations and reach the S2Cont subpopulation.

At ms 3 and 5, neuron 0 in S2Cue and neurons 0, 1, 8 and 9 in S2Cont fire due to the activity of S1Cue and S1Cont (via S2Cond), respectively. Simultaneous activation of both subpopulations leads to the activation of STDP and, thus, to the learning of the memory through content-cue associations.

Finally, S2Cue and S2Cont activity propagates to the MergeCue and MergeCont subpopulations, respectively, coinciding with the arrival of the 3-ms-delayed activity of S1Cue and S1Cont. The output activity, after the merger, will be the same as the input activity. Therefore, at ms 4 and 6, neuron 0 in MergeCue fires, as well as  neurons 0, 1, 8 and 9 in MergeCont. This activity reaches the output population at the next ms (ms 7).

The same occurs for the other two learning operations in ms 10 to 17 and ms 20 to 27 for memories formed by the activation of neurons 4, 6, 10 and 11 and neurons 3, 9, 10 and 11, respectively. After these operations, 3 memories are stored in the network by means of the associations between their cues with their contents and vice versa, i.e., with the possibility of being addressed from any of their parts.

\subsubsection{Recall}
\label{subsec:test_operations_recall}

The first recall by cue operation begins at ms 30 with the input activity of neuron 0, corresponding to the memory identified by the cue neuron 0. This activity reached S1Cue, causing the activation of neuron 0 at ms 31. This activation propagates through the synapses with STDP, activating those neurons that, due to the previous learning operation, are associated with that cue. In other words, the content of the memory is recalled from its cue. In this case, the memory presented a content formed by the activation of neurons 0, 1, 8 and 9 in S1Cont at ms 32.

Although there is activity in the S2Int neuron at ms 33, this does not occur with the S2Cond neurons, since, in this operation, they were inhibited by S1Cue. Therefore, in the second structure only neuron 0 of S2Cue was activated at ms 33. This activity, together with that of S1Cue, reached MergeCue, activating neuron 0 at ms 34. Similarly, the delayed arrival of S1Cont reached MergeCont, activating neurons 0, 1, 8 and 9 at ms 35, which corresponds to the content of the recalled memory. Finally, the recalled memory arrived at the output population (the given input cue for the operation can be seen in the output population at ms 35, and the remaining content of the memory at ms 36).

Next, a recall by content operation is initiated by passing a fragment of content as input to the network. This content fragment corresponds to the activation of neuron 11 at ms 40. This activity propagates to S1Cont at ms 41, causing the activation of neuron 6, and then propagating to the second structure. In addition to the activation of the S2Int neuron, S2Cond neuron 6 was also activated since it was not inhibited by S1Cue, both at ms 42. This leads to the activation of S2Cont neuron 6 at ms 43.

The activation of S2Cont neuron 6 propagates through STDP synapses, activating those S2Cue neurons associated with that content (due to the previous learning operation). In other words, those cues associated with at least one part of that content are recalled. In this case, this content belonged to the memories identified by S2Cue neurons 3 and 4 at ms 44. S2Cont and S2Cue activity reached MergeCont and MergeCue at ms 44 and 45, respectively, generating output activity that is identical to the input. Finally, the network recalled memories: the activity of the neurons related to the content (6) can be seen at ms 45, and the activity of the neurons related to the cues (3 and 4) at ms 46.

The recall by content operation can be performed by the activation of more than one neuron related to the content, as is the case of the operation observed in ms 50 to 56. The flow of activity is identical to that described above, but, at the time of recalling, the activation of cue neurons is the result of the sum of the recall by content operations for each of the content neurons separately.

\subsubsection{Learn with forget and recall}
\label{subsec:test_operations_forget}

The forgetting operation does not occur explicitly in the network; therefore, in order to demonstrate it, a learning operation of a memory identified by a cue belonging to a previously learned memory is performed. The previously learned memory is formed by the activity of neurons 3, 9, 10 and 11 (neuron 3 of cue and neurons 4, 5 and 6 of content), while the new memory is formed by the activity of neurons 3, 6, 8, 9 and 13 (neuron 3 of the cue and neurons 1, 3, 4 and 8 of the content). 

At ms 60, the learning operation starts by introducing the new pattern as input to the network for 3 ms. This generates the activation of neuron 3 of S1Cue and neurons 1, 3, 4 and 8 of S1Cont at ms 61 and 63. The difference with the normal learning operation comes when, at ms 62, due to the recall by cue operation, those neurons belonging to the previous memory that do not belong to the new one (5 and 6) are activated in S1Cont. The temporal sequence of activation of these neurons leads to a double activation of the STDP in favor of learning the new pattern (ms 61 and 63) and, at the same time, to an activation of the STDP for forgetting the previous pattern (ms 62).

This activity propagates to the second structure generating the activation of neuron 3 of S2Cue and neurons 1, 3, 4 and 8 of S2Cont at ms 63 and 65, as well as neurons 5 and 6 of S2Cont at ms 64. As in the first structure, the STDP mechanism is activated, learning the new memory and selectively forgetting the previous one. 

To verify whether the result of the operation was correct in both directions, a recall by cue operation (ms 70 to 76) and a recall by content operation (ms 80 to 86) were performed. The recall by cue operation was carried out with the activation of cue neuron 3, and it can be observed that the recalled content at the output corresponds to that of the new memory (neurons 1, 3, 4 and 8 of MergeCont in ms 75). The recall by content operation was performed using content neuron 6, belonging to the previous memory, but not to the new one. After performing the operation, only cue neuron 4, was activated in MergeCue (ms 85). In short, the new memory was learned, forgetting the previous one at the same time.

\subsection{Memtest86 testbenches}
\label{subsec:testbench}

\begin{figure*}[!t]
    \centering
    \includegraphics[width=0.75\textwidth]{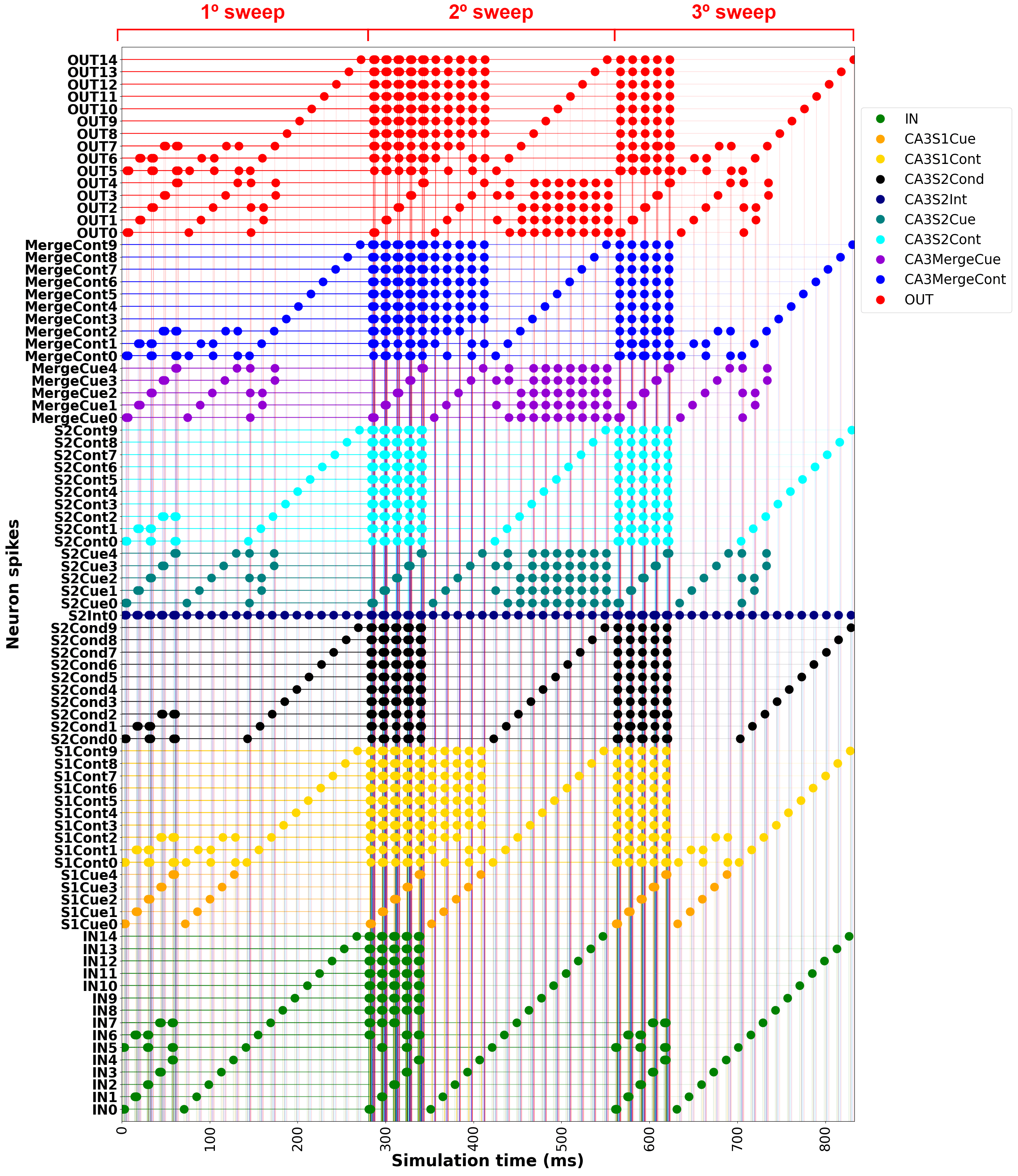}
    \caption{Raster plot of spiking activity of the network when applying the MemTest86 algorithm.}
    \label{fig:testbench}
\end{figure*}

In this experiment, a memory testbench was performed based on the MemTest86 algorithm\footnote{“MemTest86 - The standard for memory diagnostics,” \url{http://www.memtest86.com/}.}, which was adapted for this case and to also include addressing by content operations. This testbench applies an operational stress load on the memory model through 3 operation sweeps. In the first sweep, for each cue, it learns the cue itself encoded in binary as content (stores the maximum number of possible memories) and, in addition, performs one recall by cue for each possible cue and one recall by content for each possible content neuron. In the second memory sweep, the model learns in each cue the complementary of its content and checks the result with one recall by cue for each possible cue and one recall by content for each possible content neuron. Finally, in the third sweep, the same operations are performed as in the second sweep, thus the expected result is that the memory maintains the original representation of the input memories.

The test is able to use the entire memory and, at each sweep, recall operations are performed to verify whether the content is consistent with the learning operations. It was applied on the implementation of the model explained in the previous test (Section~\ref{subsec:test_operations_forget}), i.e., a model with capacity for 5 memories with a size of 15 neurons. For this implementation, the experiment took 851 ms to complete, and a total of 60 operations were performed: 15 learning operations (10 of them with forgetting), 15 recall by cue operations and 30 recall by content operations. 

The result is shown in Figure~\ref{fig:testbench}. The rasterplot is divided into the 3 sweeps, marked in red at the top, and each sweep is further divided into 2 parts: the learning operations and the recall operations.

The first sweep begins with the learning of the 5 memories formed by cue neuron \textit{i} and the encoding of \textit{i} in binary as content. For example, the memory formed by cue neuron 2 will be number 3 (starting with 1), thus the content will be formed by the binary encoding of 3, i.e., the activation of content neurons 0 and 1. On the one hand, in the recall by cue, for each cue, its binary encoding as content is obtained correctly. On the other hand, in recall by content, only the first 3 content neurons participate in memories, since 3 neurons are enough to binary encode the number of memories considered in the experiment. Thus, for example, content neuron 0 triggers the recall of cue neurons 0, 2 and 4, as it participates in the binary encoding of these 3 cues.

In the second sweep, when a complementary activity is given, the spike pattern of the figure is exactly complementary to that of the first sweep. In learning, the forgetting is not well appreciated; however, when the recall is performed, its correct operation can be observed. For example, for the memory with cue neuron 0, the original memory was formed by the activation of neuron 0, whereas for this sweep, the content is that of the activation of all content neurons except 0.

Finally, in the third sweep, the same behavior is obtained as in the first sweep, since, after complementing the content twice, the original one is obtained. The only difference is that, when performing the learning operations, in this sweep, not only the input pattern is learned, but also the previous one has to be forgotten.

\subsection{Environment state map application}
\label{subsec:test_app}

\begin{figure*}[!t]
    \centering
    \includegraphics[width=0.99\textwidth]{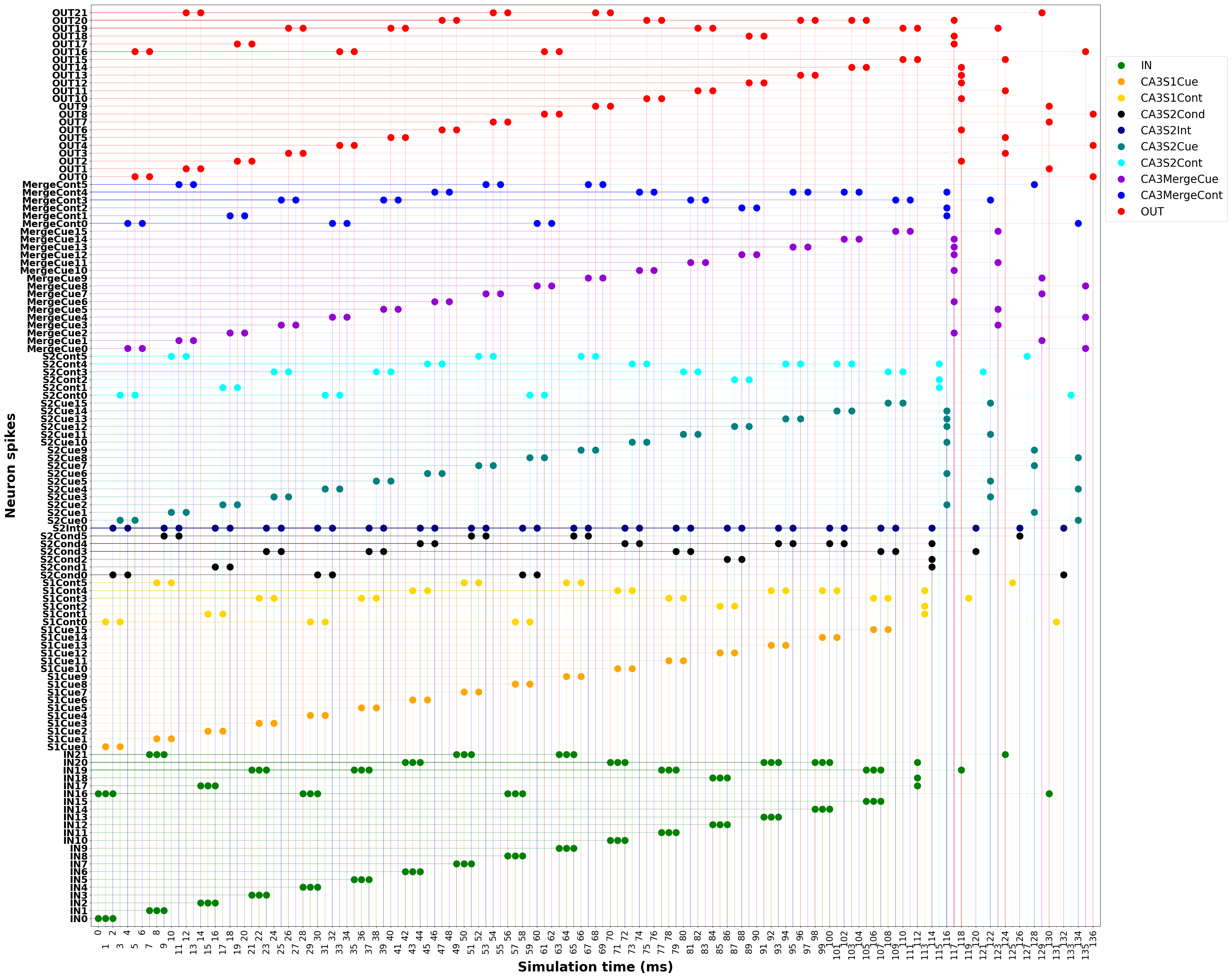}
    \caption{Raster plot of spiking activity of the network during simulation of the environment mapping application.}
    \label{fig:test_app}
\end{figure*}

\begin{figure}[!t]
    \centering
    \includegraphics[width=0.4\textwidth]{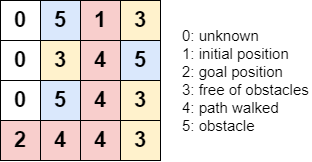}
    \caption{State map of the grid environment after reaching the robot's goal in the environment mapping application. The result of the different recall by content operations are marked in colors.}
    \label{fig:state_map}
\end{figure}

As a last experiment, an application was developed as a real use case of the model, with special emphasis on recalling memories by their content. This application extends the functionality of the one presented in the paper \cite{casanuevamorato2023bioinspired}, in which a robotic application is developed for navigation and mapping of the environment through a spiking system. The robot is located on a grid map and, for each position it goes through, it elaborates a local mapping to decide the next position to move to in order to reach its goal. After reaching the target, it maintains a spiking representation of the state of the environment it has gone through.

For this experiment, a robot within a virtual grid environment with size 4x4 was simulated. For each position, the robot is able to discern a total of 6 different states. The implementation of the model for this experiment had a storage capacity of 16 memories (one for each position of the environment) and a size of 22 neurons per memory. The memories consisted of 16 neurons encoding the cue and 6 neurons to encode the content, one for each possible state of the position within the environment: unknown (neuron 0), initial position (neuron 1), goal position (neuron 2), free of obstacles (neuron 3), visited position (neuron 4) and obstacle (neuron 5).

The result of the experiment can be seen in Figure~\ref{fig:test_app}. In the first half of the simulation (ms 0 to 112), the robot navigated through the environment, providing the state of the different positions it passed through as input to the model until reaching the goal. In the second half of the simulation (ms 112 to 136), after the state map of the environment is available (Figure~\ref{fig:state_map}), in a single recall by content operation, the system could know which positions present certain states of interest. 

At ms 112, the positions of the path taken by the robot from the initial position to the goal position, or, in other words, those positions with state 1, 2 or 4, were searched (recall by content operation using these 3 states as input). As a result, at ms 118, positions (cue neurons) 2, 6, 10, 12, 13 and 14 are obtained (marked in red in Figure~\ref{fig:state_map}). At ms 118, those positions that have no obstacles (content neuron 3) are searched to calculate new possible paths from the initial position to the goal position, obtaining positions 3, 5, 11 and 15 as a result at ms 124, marked in orange. At ms 124, positions with obstacles (content neuron 5) are searched in order to analyze the topology of the environment, obtaining positions 1, 7 and 9 at ms 130, marked in blue. Finally, at ms 130, positions with unknown state (content neuron 0) are searched to explore them in future incursions, obtaining positions 0, 4 and 8 at ms 136, marked in white.

\section{Discussion}
\label{sec:discussion}


The results of the experiments defined in Section~\ref{sec:experiments_and_results} demonstrate the correct functioning and operational capabilities of the proposed model: learning, recall by cue, recall by content and forgetting. These experiments were not limited to simple demonstrations of each individual operation, but also included the application of a memory testbench (MemTest86), widely used in the literature. This testbench took the model to a maximum stress situation, applying all types of operations at maximum frequency and making use of the maximum storage capacity available (for the implementation used).

In addition, to demonstrate its usefulness, the model was used in a simplified version of a state-of-the-art spiking robotic application. This incorporation increased the functional capabilities and performance of the system. The recall by content operation simplified memory search processes within the state map of the environment. In a single operation it was able to return all positions of interest based on their state. This operation needs 6 time steps to be performed, equal to that of the recall by cue operation, which is competitive not only in the spiking domain, but also in the state of the art of content-addressable memories if a direct hardware implementation of the model was available.


The proposed model is bio-inspired in the hippocampus, specifically in the CA3 region. The biological model is defined as a large recurrent collateral network of pyramidal neurons, but the degree and distribution of collateral connectivity of one neuron with neighboring neurons is unknown. In view of this, in this work we proposed a topological distribution of connectivity between CA3 neurons defined in Section~\ref{sec:model}.

The proposed model makes use of recurrent inhibitory collateral connections (synapses from CueCue to ContCond and from CueCont to itself) and interneurons (ContCond and ContInt) to regulate the internal activity of the network, as in the biological model. In addition, the model is able to recall a memory based on any part of it, with certain nuances to be pointed out. When recalling by content, the entire memory is not recalled, but all those cues that present at least a fragment of that content. However, the cue itself is sufficiently representative to know the complete memory by using it as input again.

Finally, regarding the encoding of the cue, in the biological model of the hippocampus, the DG region is in charge of encoding certain parts of the memory to increase its dispersion and, thus, improve its capacity to learn, store and recall memories. However, in biology, neither the encoding carried out nor the degree of dispersion achieved is known. Therefore, in view of this, it was considered that the maximum possible dispersion is carried out, i.e., a one-hot encoding on a region of the memory that we call cue.

Comparing the model described in this work with the proposals of other authors presented in Section~\ref{sec:introduction}, due to the scarcity of proposals for spiking CAM systems, bio-inspired spiking memory systems were also considered in this comparison. On the one hand, compared to the main spiking CAM proposals in the literature \cite{mueller1999content, matsugu1994spatiotemporal}, they were only able to work correctly with orthogonal patterns. The proposed model is able to work correctly with both orthogonal and non-orthogonal patterns.

On the other hand, with respect to spiking memory systems, the proposed model is bio-inspired and is purely spike-based, as opposed to \cite{zhang2016hmsnn}, which presents a hybrid system using both ANNs and SNNs, or \cite{yue2023hybrid}, which performs a direct conversion from ANNs to SNNs. The models presented in \cite{tan2011associative, tan2013hippocampus, casanueva2022spike}  are bio-inspired in the hippocampus; however, they present small storage capacities, do not work well with non-orthogonal patterns and do not allow recalling a cue by the content of a memory, while the model proposed in this work presents large adjustable storage capacities and works well with both types of patterns. Finally, the proposed model is bio-inspired in the CA3 model presented in \cite{casanueva2022bio}.
\section{Conclusions}
\label{sec:conclusions}

In this work, a fully-functional CAM model bio-inspired in the CA3 region of the hippocampus was proposed and implemented with SNNs on the SpiNNaker hardware platform. The model is capable of performing learning, recall by cue, recall by content and forgetting operations. To demonstrate the correct functioning of the model and its different operations, as well as its applicability, a set of experiments based on operational tests, stress tests using the adapted MemTest86 algorithm and a real use case based on a robotic environment mapping application were performed.

The experiments gave consistent and promising results at the functional and temporal level on two specific implementations of the model. Thanks to the parameterized architecture, it could be tested on different implementations of the model with greater or smaller size both in number of memories it can store as a maximum, and in the number of neurons for each memory. 

The proposed model was compared with its biological counterpart, pointing out its similarities and innovative proposals, as well as with other proposals in the literature. The novelty of the proposed model lies in its bio-inspired character, the possibility of learning and recalling both orthogonal and non-orthogonal memories in a consistent and stable way, and a memory recall based on any fragment of itself. In other words, a spiking, functional and bio-inspired CAM system is proposed, all this thanks to the proposed topological connectivity between the different neurons that make up CA3.

Nevertheless, the CAM system still has room for improvement or extensions that remain for future work. The hardware implementation of the spiking model and its comparison with other CAM memories in the literature would be very interesting. On the one hand, from a biological and functional point of view, the recall by content operation could be modified to return not only the cue, but the complete memory. On the other hand, another modification could be that the recall by content would return only those memories that present the exact same pattern as the one given in the input. Another interesting aspect to explore would be the applicability. An example of the possibilities of the model was demonstrated in an environment mapping application, although it could be employed in robotic swarms or hive mind architectures. In such applications, the proposed model would be able to solve problems of searching which elements have the different states of interest in a single operation, saving computational and time resources. In short, the model presents a great potential to be used in spike-based robotic applications.

The source code of the implemented model and the experiments and simulations performed is available on an open-source  GitHub repository\footnote{\url{https://github.com/dancasmor/An-aproach-to-a-spike-based-Content-Addressable-Memory-bio-inspired-in-the-Hippocampus}}.

\section*{Acknowledgments}
This research was partially supported by project PID2019-105556GB-C33 funded by MCIN/ AEI /10.13039/501100011033 and project TED2021-130825B-I00. D. C.-M. was supported by a "Formaci\'{o}n de Profesor Universitario" Scholarship from the Spanish Ministry of Education, Culture and Sport.

\printcredits

\bibliographystyle{unsrt}

\bibliography{cas-refs}




\end{document}